\documentclass[10pt,journal,compsoc]{IEEEtran}

\ifCLASSOPTIONcompsoc
  \usepackage[nocompress]{cite}
\else
  \usepackage{cite}
\fi

\usepackage{hyperref}       
\usepackage{url}            
\usepackage{booktabs}       
\usepackage{amsfonts}       
\usepackage{nicefrac}       
\usepackage{microtype}      
\usepackage{graphicx}
\usepackage{multirow}
\usepackage{enumitem}
\usepackage{subfig}

\usepackage{amsmath}
\usepackage{enumitem}
\usepackage{booktabs}
\usepackage{hyperref}
\usepackage{tabularx}

\usepackage{soul}
\soulregister\cite7 
\soulregister\ref7 

\usepackage{adjustbox}
\usepackage{bm}
\usepackage{balance}
\usepackage{dblfloatfix}
\usepackage[vlined,linesnumbered,titlenumbered,ruled]{algorithm2e}
\usepackage{array, boldline, makecell, booktabs} 

\usepackage{tabularx}{\tiny}
\usepackage{color, colortbl}
\usepackage{graphicx}
\usepackage{epsfig}
\usepackage{mathrsfs}
\usepackage{multirow}
\usepackage[skip=2pt]{caption}
\usepackage{lipsum}
\usepackage{array}
\usepackage{enumitem,kantlipsum}
\usepackage{url}

\usepackage{makecell}

\begin{document}

\title{FlexSSL : A Generic and Efficient Framework for Semi-Supervised Learning}

\author{Huiling Qin, Xianyuan Zhan, Yuanxun Li, Yu Zheng
\IEEEcompsocitemizethanks{
  \IEEEcompsocthanksitem Huiling Qin, Yu Zheng are with Xidian University and JD iCity, JD Technology, Beijing, China; JD Intelligent Cities Research. E-mail: orekinana@gmail.com and msyuzheng@outlook.com \protect
  \IEEEcompsocthanksitem Yuanxun Li is with JD.COM, Beijing, China. E-mail: genkunabe@gmail.com \protect
  \IEEEcompsocthanksitem Xianyuan Zhan is with Institute for AI Industry Research (AIR), Tsinghua University, Beijing, China. E-mail: zhanxianyuan@gmail.com
}
}


\IEEEtitleabstractindextext{%
\begin{abstract}

	Semi-supervised learning holds great promise for many real-world applications, due to its ability to leverage both unlabeled and expensive labeled data. However, most semi-supervised learning algorithms still heavily rely on the limited labeled data to infer and utilize the hidden information from unlabeled data. We note that any semi-supervised learning task under the self-training paradigm also hides an auxiliary task of discriminating label observability. Jointly solving these two tasks allows full utilization of information from both labeled and unlabeled data, thus alleviating the problem of over-reliance on labeled data. This naturally leads to a new generic and efficient learning framework without the reliance on any domain-specific information, which we call FlexSSL. The key idea of FlexSSL is to construct a semi-cooperative ``game”, which forges cooperation between a main self-interested semi-supervised learning task and a companion task that infers label observability to facilitate main task training. We show with theoretical derivation of its connection to loss re-weighting on noisy labels. Through evaluations on a diverse range of tasks, we demonstrate that FlexSSL can consistently enhance the performance of semi-supervised learning algorithms.
\end{abstract}

\begin{IEEEkeywords}
Semi-Supervised Learning, Deep Learning, Pseudo Labeling.
\end{IEEEkeywords}}

\maketitle

\IEEEdisplaynontitleabstractindextext

\IEEEpeerreviewmaketitle

\ifCLASSOPTIONcompsoc
\IEEEraisesectionheading{\section{Introduction}\label{sec:introduction}}
\else
\section{Introduction}
\label{sec:introduction}
\fi
Despite the huge success of supervised learning with deep neural networks,
they require large amounts of labeled data for model training to achieve high performance \cite{DBLP:conf/aaai/MengSZH19, DBLP:conf/kdd/PrabhuV14}. Collecting labeled data can be very difficult and costly in practice \cite{DBLP:conf/emnlp/MengZHXJZH20}. The challenge can be further exacerbated as in many cases, human or machine labeled data may contain errors or noises \cite{DBLP:journals/jmlr/NatarajanDRT17, DBLP:conf/icml/SchnabelSSCJ16}. Semi-Supervised Learning (SSL)
\cite{DBLP:books/mit/06/CSZ2006}
that leverages both labeled and unlabeled data has become an increasingly promising yet still challenging area, and is widely applicable to many real-world problems.

To utilize the hidden information in the unlabeled data, a general and popular semi-supervised learning paradigm is through self-training \cite{yarowsky1995unsupervised,lee2013pseudo}.
Self-training uses a previously learned model to predict labels for the unlabeled data (pseudo-labeling) which are then used for subsequent model training \cite{yarowsky1995unsupervised,lee2013pseudo,DBLP:conf/iclr/LaineA17,tarvainen2017mean,iscen2019label}. Despite some empirical successes, self-training methods still suffer from two core challenges, which are \textit{over-reliance on labeled data} and \textit{error accumulation}. Most semi-supervised learning methods \cite{yarowsky1995unsupervised,lee2013pseudo,DBLP:conf/iclr/LaineA17, DBLP:conf/nips/SohnBCZZRCKL20, DBLP:conf/cvpr/XieLHL20, DBLP:conf/iclr/KipfW17} assume the labeled and unlabeled data follow the same or similar data distribution, thus the data-label mapping mechanisms learned from the labeled data are somewhat transferable to unlabeled data. Essentially, the algorithm is still reinforcing the information in the labeled data rather than mining additional information from the unlabeled data. Second, the pseudo-labels of unlabeled data can be incorrectly predicted. Using such biased information for training in subsequent epochs could increase confidence in erroneous predictions, and eventually leading to a vicious circle of error accumulation \cite{cai2013heterogeneous,arazo2020pseudo}. The situation can be even worse when the labeled data contain errors or noises, as the model learned from labeled data will make more mistakes, resulting in more severe error accumulation.

Alleviating the issue of over-reliance on labeled data requires extracting additional sources of information from the unlabeled data. Some recent approaches introduce domain-specific data augmentation and consistency regularization \cite{sajjadi2016regularization} to enforce the stability of the predictions with respect to the transformations of the unlabeled data (e.g. data augmentation on images such as rotation, flip and shift, etc.).
Although such data perturbation or augmentation are well-defined for images, they are not directly transferable to general settings due to strong reliance on the domain-specific knowledge, hence the success of these methods are mostly restricted to image classification tasks \cite{sajjadi2016regularization,liu2019decoupled,DBLP:conf/cvpr/XieLHL20,DBLP:conf/nips/SohnBCZZRCKL20}. 
We provide a new insight by noting that, under the self-training paradigm, every semi-supervised learning task hides another auxiliary task of discriminating whether the data label is real or a pseudo-label predicted by a machine oracle. Jointly solving the main task together with the auxiliary task allows sufficient utilization of the hidden information in unlabeled data without the need of any domain-specific information.

To break the vicious cycle of error accumulation, one needs to not always trust the machine-labeled data. A common approach in semi-supervised learning literature is to use stricter and complex "rules" for selecting high-quality pseudo-labels, but at the cost of extra high computation burden during the label selection process~\cite{DBLP:conf/iclr/RizveDRS21,zhang2021flexmatch}. 
An alternative idea is to incorporate the reliability or confidence of data labels and treat them differently during training. 
A connected approach in supervised learning is to use the notion of "soft labels" with confidence probabilities or weights, which have been validated in a number of noisy label learning and pseudo-labeling problems \cite{DBLP:journals/jmlr/NatarajanDRT17, DBLP:conf/icml/SchnabelSSCJ16,DBLP:conf/aaai/Shen0ZK20, algan2021meta}. This treatment can also help semi-supervised learning algorithms to gain stable learning from noisy labels.

Combining previous insights, we develop a new generic and efficient semi-supervised learning framework, called FlexSSL. FlexSSL contains three ingredients. First, it simultaneously solves a companion task by learning a discriminator as a critic to predict label observability (whether it is a true label or a pseudo-label), which is used to mine additional information in the unlabeled data while facilitating main task learning. Second, the output of the discriminator is in fact a measure of label reliability, thus can be interpreted as the confidence probabilities of labels. This converts the original problem into a soft label learning problem. Last but not least, FlexSSL introduces a cooperative-yet-competitive learning scheme to boost the performance of both the main task and the companion task. 
Specifically, FlexSSL forges cooperation between both tasks by providing extra information to each other (the main task provides prediction loss, companion task provides label confidence measures), leading to improved performance of both tasks. Moreover, we consider the main task model to be self-interested which tries to challenge the discriminator by providing as little information as possible. This design forms a semi-cooperative ``game'' with a partially adversarial main task model, which can be shown equivalent to a loss reweighting mechanism on noisy soft labels. 

The proposed FlexSSL can be easily incorporated into a wide range of SSL models (e.g. image classification, label propagation and data imputation) with limited changes or extra computation cost on the original method. We show with empirical experiments that FlexSSL consistently achieves superior performance and effectiveness compared with the original and self-training version of different SSL algorithms. Moreover, FlexSSL naturally provides the data confidence measures from the discriminator as a byproduct of the learning process, which offers additional interpretability of the semi-supervised learning task. This can be particularly useful for many practical applications, while also providing extra benefits for scenarios with noisy data labels.

\section{Preliminaries}

\subsection{Problem Statement}
We formulate the semi-supervised learning problem as learning a model $f(\cdot)$ with input data $x\in X$ to predict the label $y\in Y$. In SSL, only a partial set of data labels $Y_L$ with ground truth are given in the training set, the rest are unlabeled $Y_U$ ($Y = Y_L \bigcup Y_U$). For convenience of later discussion, we denote $L$ as the set of indices of data samples with labels and $U$ as the set of indices for unlabeled data.
In many real-world SSL problems, the size of labeled data $|L|$ is often limited. In extreme cases, $Y_L$ may potentially contain errors or noises.
We further define a mask vector $M$ to denote the observability of labels over each data sample. We set $M_i \in M$ equals to 1 if $i \in L$ and $M_j \in M$ equals to 0 if $j \in U$.
To develop a generic SSL framework, we consider a diverse set of task settings, include:

\begin{itemize}[leftmargin=*,nosep]
   \item \textbf{Inductive classification:} a typical SSL task setting is to construct a classifier to predict labels $y\in Y$ for any object in the input space $x\in X$. Common examples include semi-supervised image classification tasks \cite{sajjadi2016regularization,DBLP:conf/nips/SohnBCZZRCKL20,DBLP:conf/cvpr/XieLHL20}, where only a subset of data samples are given the known labels $Y_L$ and the rest data labels $Y_U$ are unknown. 
   \item \textbf{Label propagation:} another class of SSL tasks under transductive setup is to use all input samples $X$ and observed labels $Y_L$ to train a classifier to predict on unseen labels $Y_U$. Examples include classifying nodes in a graph given only small subset of node labels $Y_L$ \cite{DBLP:conf/iclr/KipfW17}.
   \item \textbf{Data imputation:} a special SSL task under regression setting, in which the missing state of input and output data are strongly correlated \cite{qin2021network,DBLP:conf/cvpr/RichardsonWLXB20}. In data imputation, the inputs $X$ are partially filled and the labels $Y$ are equivalent to a reconstructed version of $X$ obtained through regression.
\end{itemize}

\subsection{Auxiliary Task in SSL Problems}
We begin our discussion by first noting that, if we feed the unknown labels $Y_U$ with the model $f$ predicted labels $\Tilde{Y}_U=\hat{Y}_U^{(k)}$ ($\hat{Y}_U^{(k)}=f(X)$ represents the labels from the $k$-th round of pseudo-labeling) or other machine-generated labels, there actually hides an auxiliary task of discriminating whether the data label is real or a pseudo-label. 
Denote all the data labels under pseudo-labeling as $\Tilde{Y}=Y_L\bigcup \Tilde{Y}_U$. 
We can train a discriminator $d(\cdot)$ to tell the confidence measure $p\in P$ of whether $x\rightarrow \Tilde{y}$ ($x\in X, \Tilde{y}\in \Tilde{Y}$) is a valid mapping. This is always learnable since the ground truth of $P$ is exactly the observability mask $M$. 
With a slight abuse of notation, we formulate the main SSL task A and the auxiliary companion task B as follows:
\begin{gather}\label{eq:definition}
    \begin{aligned}
    &A: f(X)=\hat{Y}, &\mathcal{L}_A = loss_A(Y_L, \hat{Y}_L)\\
    &B: d(X, \Tilde{Y})=P, &\mathcal{L}_B = loss_B(P, M)
     \end{aligned}
\end{gather}
where $loss_A$ is the original loss function of the main task and $loss_B$ can be any binary classification loss function between $P$ and $M$, such as the binary cross entropy (BCE) loss. Jointly solving the above two tasks allows exploiting the underlying relationship between input data $X$ and data label $Y$ from another angle, which can potentially provide more information to facilitate main task training.

\begin{figure*}[tb]
  \centering
  \includegraphics[width=0.98\textwidth]{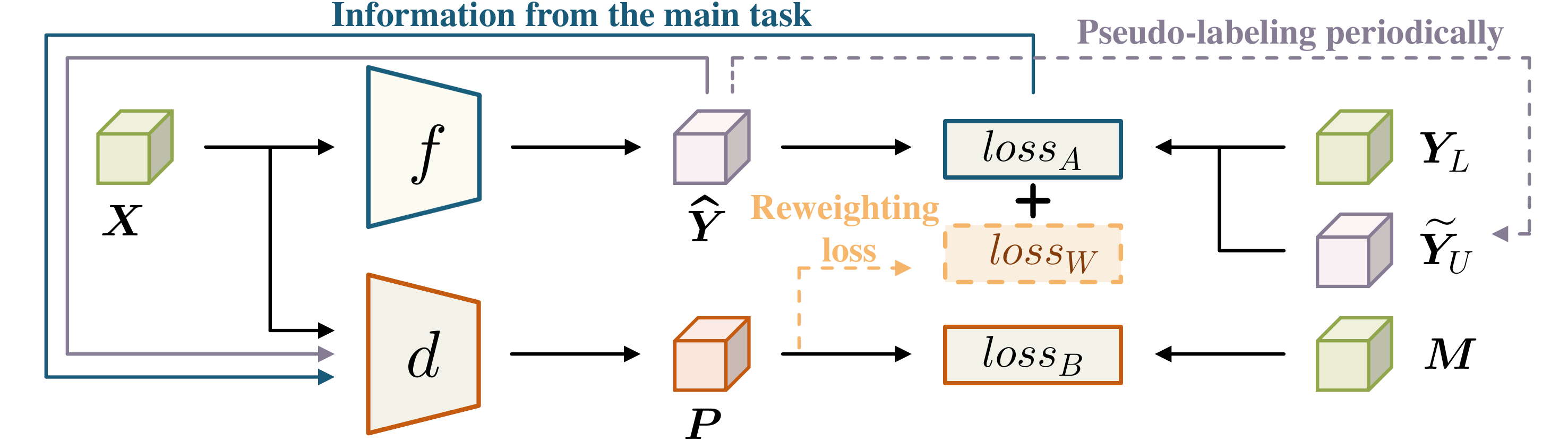}
  \caption{Illustration of the learning strategy in FlexSSL}
  \label{fig:framework}
\end{figure*}

\section{The FlexSSL Framework}
Note that the two tasks specified in Problem (\ref{eq:definition}) are independent with each other, it is not possible to gain much learning benefit unless we introduce some form of coupling between the main model $f$ and the discriminator $d$. In this section, we introduce FlexSSL to jointly learn $f$ and $d$ in a principled way.
It forges cooperation between a self-interested main task and an unselfish companion task for boosted performance. We show by theoretical derivation that the introduction of auxiliary tasks under the pseudo-labeling paradigm will lead to a soft-label learning formulation of the original SSL task. 

\subsection{Alternative Formulation Under FlexSSL}
We consider an alternative formulation of Problem (\ref{eq:definition}) by modifying the loss of main task $\mathcal{L}_A$ and inputs of the companion task to forge cooperation and information sharing:
\begin{align}
    \label{eq:SCL_definition}
    &A: f(X)= \hat{Y},\;\mathcal{L}_A = loss_A(\Tilde{Y},\hat{Y}) + \alpha \cdot loss_W(\Tilde{Y},\hat{Y}, P) \notag\\
    &B: d(X, \hat{Y}, g(\Tilde{Y}, \hat{Y})) = P,\quad  \mathcal{L}_B = loss_B(P, M) 
\end{align}    
where $g(\Tilde{y}, \hat{y})$ is some form of information provided by the main task.
We can simply model the information $g(\cdot)$ as the element-wise loss term of the main task, i.e. $g(\Tilde{y}, \hat{y})=loss_A(\Tilde{y}, \hat{y})$, such as the cross entropy loss for classification tasks and mean square error (MSE) loss for regression tasks. Moreover, $loss_W(\Tilde{Y},\hat{Y}, P)$ is a corrective additional loss term impacted by the label confidence measure information $P$ provided by the discriminator $d$, and $\alpha$ is its weight parameter. Due to the existence of corrective information from $loss_W$, we now train the main task with all data labels $\Tilde{Y}$ rather than observed labels $Y_L$, thus allows sufficient utilization of information from the unlabeled data. 
Figure \ref{fig:framework} illustrates the learning strategy of FlexSSL. 

The design of FlexSSL has several interesting properties. First, since additional information are cooperatively provided by both tasks to each other, they can be both learned better.
Second, a less well-learned model $f$ will produce large element-wise error $loss_A(\Tilde{y}, \hat{y})$ for a predicted label $\hat{y}$, which in turn provides more information in $g(\cdot)$ for the discriminator $d$ to distinguish whether a label is real or a pseudo-label. As $f$ is learned better during the cooperative learning process, when $g$ no longer provides useful information for discriminating the label observability, then we may have reason to believe $f$ has achieved satisfactory inference performance. Hence minimizing the information in $g(\cdot)$ coincides with improving the learning of $f$. This suggests that if we alter the learning direction of $f$ using $loss_W$ to encourage it providing as little information in $g(\cdot)$ as possible for the discriminator $d$, then we might obtain a better learned $f$. This in fact forms a cooperative-yet-competitive ``game'' between both tasks, with the main task being self-interested and in some sense partially adversarial to the discriminator $d$.
Under this design, the main task and companion task in FlexSSL are neither fully cooperative as in multi-objective optimization \cite{deb2014multi} nor fully adversarial to each other as in adversarial learning \cite{goodfellow2014generative,DBLP:conf/nips/MatyaskoC18}, but balance the benefit of cooperation and competition.
Finally, it also allows the discriminator to detect errors or noises in observed data labels. As the erroneous $(x, y)$ data pairs may possess different $x\rightarrow y$ mapping patterns against normal data, thus are more likely to be identified as pseudo-labels. Such negative impact can be further corrected using $loss_W$ with label confidence measure $P$.

\subsection{Derivation of the Additional Loss Term $loss_W$}
Until this point, the only mystery about FlexSSL is the exact form of $loss_W$ that embodies the self-interested behavior for the main task model $f$. We resort to calculus of variations to provide theoretical derivation of $loss_W$. 

Under the alternative formulation (\ref{eq:SCL_definition}), both the discriminator $d$ and its loss function $loss_B$ are impacted by the information $g(\cdot)$ provided by $f$, thus they are now functionals of $f$ (i.e. function of a function). For simplicity, we can express $d$ and $loss_B$ as $d(x,g(f))$ and $loss_B(x,d,g(f))$. Note that under empirical risk minimization, we can write $\mathcal{L}_B$ as
where $p_{data}(x)$ is the distribution of training dataset $\mathcal{D}$ and $\Omega_\mathcal{D}$ is its domain. Observe that $\mathcal{L}_B$ has a natural integral form of the functional $F(x,d,f)=p_{data}(x) \cdot loss_B(x,d,g(f))$, which is commonly studied in functional analysis. We can thus define $\mathcal{L}_B$ as a new functional $J(f,d)$.

Although the discriminator $d$ aims to minimize its loss $\mathcal{L}_B$ (or $J(d,f)$), as mentioned previously, we want the main task model $f$ to challenge the discriminator $d$ by providing as little information in $g$ as possible. This essentially requires changing the behavior of $f$ to be adversarial to $d$ such that $J(d,f)$ attains its maxima. This results in solving following minimax optimization problem for $J(d,f)$:
\begin{gather}
  \label{eq:minimax_definition}
  \min_d \max_f J(d,f)
\end{gather}
We wish to examine how the variation of $f$ impacts $J(d,f)$. 
To simplify the analysis, we first focus on solving the inner maximization problem with respect to $f$, and consider $d$ as an unknown external functional decided by the outer minimization problem. 

According to calculus of variation, the extrema (maxima or minima) of a functional can be obtained by  
solving the associated Euler-Lagrange equation: $F_f - \frac{\text{d}}{\text{d}x}F_{f'} = 0$,
where $F_f=\partial F / \partial f$ and $F_{f'}=\partial F / \partial f'$. In our case $\frac{\text{d}}{\text{d}x}F_{f'} = 0$ as $f'$ does not appear in functional $J(d,f)$. As $F$ is a function of $d$, and $d$ is a function of $g(f)$, therefore, 
it is necessary that $F_f = \frac{\partial F}{\partial d}\cdot \frac{\partial d}{\partial g}\cdot \frac{\partial g}{\partial f} = 0$ holds.
Let $\theta_f$ be the model parameters of $f$, it also suggests that
\begin{gather}
\begin{aligned}\label{eq_product}
  \frac{\partial F}{\partial d}\cdot \frac{\partial d}{\partial g}\cdot \frac{\partial g}{\partial f}\cdot \frac{\partial f}{\partial \theta_f}  &\overset{(i)}{=} \frac{\partial F}{\partial d}\cdot \frac{\partial d}{\partial g}\cdot \nabla_{\theta_f} loss_A 
  \\ &\overset{(ii)}{=} \frac{\partial d}{\partial g}\cdot \textcolor{red}{\frac{\partial F}{\partial d}\cdot \nabla_{\theta_f} loss_A} = 0
  \end{aligned}
\end{gather}

The first equation $(i)$ is due to our design of letting $g(\cdot)$ be the element-wise loss function of $f$ ($g(\Tilde{y}, \hat{y})=loss_A(\Tilde{y}, \hat{y})$). For the second equation $(ii)$, as $d$, $g$ and $F$ are real-valued functionals of $x$ and $f$, which is the same for their partial derivatives $\partial F/\partial d$ and $\partial d/\partial g$. Assume the continuity of both $\partial F/\partial d$ and $\partial d/\partial g$ are satisfied, as the set of real-valued continuous functions is a commutative ring \cite{hewitt1948rings}, hence we can swap their order for convenience.

As $d$ is determined by the outer minimization problem of (\ref{eq:minimax_definition}), thus $d$ is unkown and it is
not possible to obtain the exact $\partial d/\partial g$ by only inspecting the inner maximization problem. Instead, we consider another solution of Eq.(\ref{eq_product}) by letting $\frac{\partial F}{\partial d}\cdot \nabla_{\theta_f} loss_A = 0$. Directly solving this functional equation for $\forall x\in \mathcal{D}$ is still intractable, as $\frac{\partial F}{\partial d}$ and  $\nabla_{\theta_f} loss_A$ can be complicated functions and the $p_{data}(x)$ in $F$ is typically unknown. However, we can consider a relaxed and tractable condition by computing the its integration over $\Omega_\mathcal{D}$. As $\Omega_\mathcal{D}$ is bounded when $\mathcal{D}$ is finite, thus the final integral is still 0, 
\begin{align}\label{eq:condition}
  0 &= \int_{\Omega_\mathcal{D}} \frac{\partial F}{\partial d}\cdot \nabla_{\theta_f} loss_A \text{d}x \notag\\
  &= \int_{\Omega_\mathcal{D}} p_{data}(x)\cdot \frac{\partial loss_B(x,d,f)}{\partial d} \cdot \nabla_{\theta_f} loss_A \text{d}x \notag \\
  & \approx \sum_{x\in\mathcal{D}} \frac{\partial loss_B(x,d,f)}{\partial d}\nabla_{\theta_f} loss_A
\end{align}

As an example, suppose we use BCE loss for $loss_B$, and write $p_i$ as the output value of $d(X,\hat{Y},g(\Tilde{Y}, \hat{Y}))$ with index $i$, the condition (\ref{eq:condition}) becomes:
\begin{gather}\label{eq:L_f}
	\begin{aligned}
		&-\sum_{i \in L} \frac{1}{p_i}\nabla_{\theta_f} loss_A(\Tilde{y}_i, \hat{y}_i)
		\\ &+ \sum_{j \in U} \frac{1}{1-p_j}\nabla_{\theta_f}loss_A(\Tilde{y}_j, \hat{y}_j) = 0
	\end{aligned}
\end{gather}


Note that the LHS of above equation can be perceived as the gradient of a new term $L_f(\Tilde{Y},\hat{Y}, P)$:
\begin{gather}\label{eq:L_f}
\begin{aligned}
   -L_f(\Tilde{Y},\hat{Y}, P)=&\sum_{i \in L} \frac{1}{p_i} \cdot loss_A(\Tilde{y}_i, \hat{y}_i)  
   \\ & - \sum_{j \in U} \frac{1}{1-p_j} \cdot loss_A(\Tilde{y}_j, \hat{y}_j)
\end{aligned}
\end{gather}

Consequently, minimizing $-L_f$ with respect to $\theta_f$ ($\frac{\partial L}{\partial \theta_f}=0$) satisfies the condition (\ref{eq:condition}), which is derived from the necessary condition of solving the inner maximization problem of $J(d,f)$ specified in (\ref{eq:minimax_definition}). The negative sign before $-L_f$ is to ensure gradient ascent update in the direction of $F_f$ to maximize $J(d,f)$.
We now uncover the exact form of $loss_W$, which can be set as $loss_W(\Tilde{Y},\hat{Y}, P)=-\mathcal{L}_f(\Tilde{Y},\hat{Y}, P)$ to realize the self-interested behavior of $f$. The resulting $loss_W$ can be considered as an additional reweighting loss that complements the original loss term $loss_A$ of the main task model $f$ based on the estimated confidence measures $P$ from the discriminator $d$.

\subsection{Interpretation of FlexSSL}
Adding $loss_W$ to the original loss term $loss_A$, we can obtain the complete loss for the main task as:
\begin{gather}\label{eq:loss_W}
\begin{aligned}
   \mathcal{L}_A =  &\sum_{i \in L} \Big(1+\frac{\alpha}{p_i}\Big) \cdot loss_A(y_i,\hat{y_i}) 
   \\ &+ \sum_{j \in U} \Big(1-\frac{\alpha}{1-p_j}\Big) \cdot loss_A(\Tilde{y}_j, \hat{y_j})
\end{aligned}
\end{gather}
This essentially transforms the original main task into a cost-sensitive learning problem \cite{zadrozny2003cost} by imposing following soft-labeling weights on the errors of each data label as: 
\begin{gather}\label{eq:reweight}
\scalebox{0.97}{$
   \text{Soft-labeling weights} = \left\{
  \begin{array}{lr}
  1 + \alpha/p_i, &i \in L \\
  1-\alpha/(1- p_j), &j \in U
  \end{array}
  \right.
  $}
\end{gather}

Above soft-labeling weights induce very different behaviors on the loss $loss_A(\Tilde{y}, \hat{y})$ of labeled and pseudo-labeled samples. As $p_i$ represents the judgment of the discriminator $d$ on whether $y_i$ is labeled or not. For labeled samples, the loss are boosted by additional weight of $\alpha/p_i$, which means if the discriminator $d$ judges an labeled samples $y_i\in Y_L$ to be unlabeled or unreliable (confidence $p_i\rightarrow  0$), the loss of this sample will get significantly boosted. This forces the main task model $f$ to pay more attention to infer the problematic labeled samples. As unlabeled samples $y_j\in Y_U$ are periodically filled with pseudo-labels generated by $f$ (initially filled with random values), FlexSSL does not fully trust these labels and may even choose to enlarge the gap with the pseudo-label when $p>1-\alpha$ ($\alpha \leq 1$). Specifically, the more likely an label $y_j$ is considered reliable by the discriminator ($p_j \rightarrow 1$), the stronger it forces the learning process to ignore further improvement on inferring $y_j$. 

\begin{table*}[t]
    \centering
    \begin{adjustbox}{center}
        \begin{tabular}{lccc}
           \toprule
            Discriminator loss& Loss function & 
             Weights for $i\in L$ & Weights for $j\in U$  \\ 
           \midrule
           BCE loss & $-\sum_{i \in L}\log p_i - \sum_{j \in U}\log (1-p_j)$ &  $1+\alpha/p_i$ & $1-\frac{\alpha}{1-p_j}$\\ 
           Exponential loss & $\sum_{i \in L}e^{-p_i} + \sum_{j \in U}e^{p_j}$ & $1+\alpha e^{-p_i}$ & $1-\alpha e^{p_j}$ \\
           Logistic loss  & $\sum_{i \in L}\log (1+e^{-p_i})+\sum_{j \in U}\log (1+e^{p_j})$ & $1+ \frac{\alpha e^{-p_i}}{1+e^{-p_i}}$ & $1- \frac{\alpha e^{p_j}}{1+e^{p_j}}$\\
           \bottomrule
        \end{tabular} 
        \end{adjustbox}
    \caption{Soft-labeling weights under different forms of discriminator loss functions
    }
    \label{tab:weights_form}
  \end{table*}

Note that FlexSSL is a very general framework, one can use different choices of loss function $loss_B$ for the discriminator, which may likely give rise to different forms of soft-labeling weights. In Table \ref{tab:weights_form}, we present the soft-labeling weights (with $p_i$ represents $d(x_i,g(f(x_i)))$ for simplicity) of three commonly used classification loss functions for the discriminator loss ($loss_B$) derived under the FlexSSL framework. Since the continuity of $\partial loss_B/\partial  d$ needs to be satisfied in the derivation of Eq. (\ref{eq_product}). The derivative of BCE loss can potentially result in $1/p_i$ and $1/(1-p_i)$ with infinite values, violating the continuity assumption. We thus clip their values as $\min\{1/p_i, H\}$ and $\min\{1/(1-p_i), H\}$ in our practical algorithm, where $H$ is set to $10$ in our implementation.
The same clip operation is also applicable under other loss terms.

\begin{table*}[t]  
  \centering
  \begin{adjustbox}{center}
      \begin{tabular}{@{}cc{c}ccc{c}ccc{c}cc@{}}\toprule
      \multirow{3}{*}{\begin{tabular}[c]{@{}c@{}}Missing\\ rate (\%)\end{tabular}} && \multicolumn{2}{c}{Image classification} &  & \multicolumn{2}{c}{Label propagation} & & \multicolumn{2}{c}{Data imputation}\\
      && \multicolumn{2}{c}{Accuracy (\%)} & & \multicolumn{2}{c}{Accuracy (\%)} & & \multicolumn{2}{c}{MSE}\\
      \cmidrule{3-4} \cmidrule{6-7} \cmidrule{9-10}
            && ResNet18 & ResNet18+FlexSSL && GCN   & GCN+FlexSSL && DAE & DAE+FlexSSL \\ \midrule
         00 && 83.85 $\pm$ 0.12   & 86.23 $\pm$ 0.13  && -     & -       && -   & -  \\
         10 && 83.57 $\pm$ 0.12   & 84.69 $\pm$ 0.08  && 82.18 $\pm$ 1.59 & 84.33 $\pm$ 0.56  && 0.8055& 0.7899 \\
         20 && 83.06 $\pm$ 0.16   & 85.15 $\pm$ 0.13  && 82.18 $\pm$ 1.69 & 84.00 $\pm$ 1.81  && 0.9029  & 0.8716  \\
         30 && 82.89 $\pm$ 0.21   & 85.38 $\pm$ 0.16  && 81.80 $\pm$ 1.75 & 83.33 $\pm$ 0.68  && 0.9335 & 0.8867 \\
         40 && 82.43 $\pm$ 0.13   & 84.98 $\pm$ 0.21  && 82.28 $\pm$ 2.69 & 83.00 $\pm$ 0.95  && 0.9761  & 0.9229 \\
         50 && 81.94 $\pm$ 0.16   & 84.61 $\pm$ 0.21  && 81.93 $\pm$ 2.56 & 81.33 $\pm$ 2.28  && 0.9726  & 0.9412 \\
         60 && 81.23 $\pm$ 0.16   & 83.87 $\pm$ 0.20  && 80.03 $\pm$ 6.37 & 82.33 $\pm$ 2.10  && 0.9787 & 0.9727  \\
         70 && 79.70 $\pm$ 0.13   & 82.90 $\pm$ 0.21  && 77.84 $\pm$ 7.52 & 83.67 $\pm$ 3.73  && 1.1593 & 1.1582 \\
         80 && 78.84 $\pm$ 0.35   & 81.31 $\pm$ 0.42  && 74.33 $\pm$ 7.00 & 80.33 $\pm$ 1.00  && 1.1429 & 1.1425  \\
         90 && 76.02 $\pm$ 0.41   & 77.96 $\pm$ 0.47  && 69.66 $\pm$ 7.66 & 74.26 $\pm$ 0.53  && -      &   \\
      \bottomrule
      \end{tabular}
    \end{adjustbox}
      \caption{Comparison of the original models and the FlexSSL-integrated version for different tasks (with mean $\pm$ std). The std results in the imputation task are omitted because they are less than 1e-04. Results averaged over 10 random seeds.}
  \label{tab:acc}

\end{table*}

\section{Experiments}
    FlexSSL can be easily incorporated in a variety of SSL tasks. We choose three representative SSL tasks and apply FlexSSL on their official implementations to evaluate the performance of FlexSSL, including: image classification \cite{he2016deep} for inductive classification, label propagation \cite{DBLP:conf/iclr/KipfW17} for transductive classification and data imputation \cite{DBLP:conf/pakdd/GondaraW18} for regression. We also conduct additional experiments that compare FlexSSL with some advanced SSL methods on the image classification tasks to evaluate the efficiency and effectiveness of FlexSSL.
    
     \begin{itemize}[leftmargin=*]
    \item \textbf{Image classification:}
    We use RestNet18 for image classification on the fashion-mnist dataset \cite{DBLP:journals/corr/abs-1708-07747}, with 6,000/10,000 training/testing set partition. The training set only contains 10\% of the original dataset which aims to increase the task difficulty.
    \item \textbf{Label propagation:}
    We trained a two-layer GCN \cite{DBLP:conf/iclr/KipfW17} integrated with FlexSSL and evaluate the prediction accuracy on Cora dataset \cite{sen2008collective}. The test set consists of 1000 labeled examples extracted from a knowledge graph with 2,708 nodes and 5,429 edges.
    \item \textbf{Data imputation:}
    We use a deep denoising autoencoder (DAE) \cite{DBLP:conf/pakdd/GondaraW18} for data imputation under regression setting. We use the UCI online news popularity dataset \cite{asuncion2007uci} for evaluation. The data is split the training/testing set with ratio of 0.8/0.2.
        \end{itemize}
    
    For all tasks, we remove certain proportion (\textit{missing rate}) of data labels in the training set and treat them as unlabeled data in the semi-supervised setting. 
    At initial step of FlexSSL, as the model is not yet trained to provide pseudo-labels, we fill the initial pseudo-labels for unlabeled data with random labels.
    For the regression task (data imputation), we compute the mean $\mu$ of labeled data and fill the missing entries with $\mu$ adding random noises; for the classification tasks, we fill the missing label $Y_U$ by randomly selecting a data label. 
    
    \subsection{Performance Evaluation Across Diverse Tasks}
    \subsubsection{Performance enhancement. }
    We compare in Table \ref{tab:acc} our method with the original model in three representative SSL tasks to demonstrate the flexibility and accuracy of FlexSSL under diverse problem settings. 
    FlexSSL consistently achieves superior performance over all tasks, which shows the universality of FlexSSL under diverse SSL problem setting. We also use 10 different levels of missing rates to verify the robustness of FlexSSL. By randomly dropping some labels, we increase the missing rate of the training set from 0\% (fully supervised) to 90\%. As label propagation and data imputation tasks are not well-defined under fully supervised setting, only the image classification task under 0\% missing rates is presented in the table. It is observed in Table \ref{tab:acc} that most of the original models experience substantial performance deterioration with the increase of missing rate. By contrast, FlexSSL remains high level of robustness with limited performance drop in most tasks.
    
    Surprisingly, for inductive image classification task, we observe that in the case of 0\% missing rate (fully supervised), FlexSSL can achieve even higher accuracy compared with the fully supervised base model ($86.2\pm 0.13\%$ vs $83.85\pm 0.13\%$). Moreover, FlexSSL achieves similar level of accuracy as the fully supervised base model even with $60\%$ missing rate. These results suggests that FlexSSL can also be used in supervised learning problems to boost model performance.
    
  \begin{figure*}[h]
    \centering
    \makebox[\textwidth][c]{\includegraphics[width=0.98\textwidth]{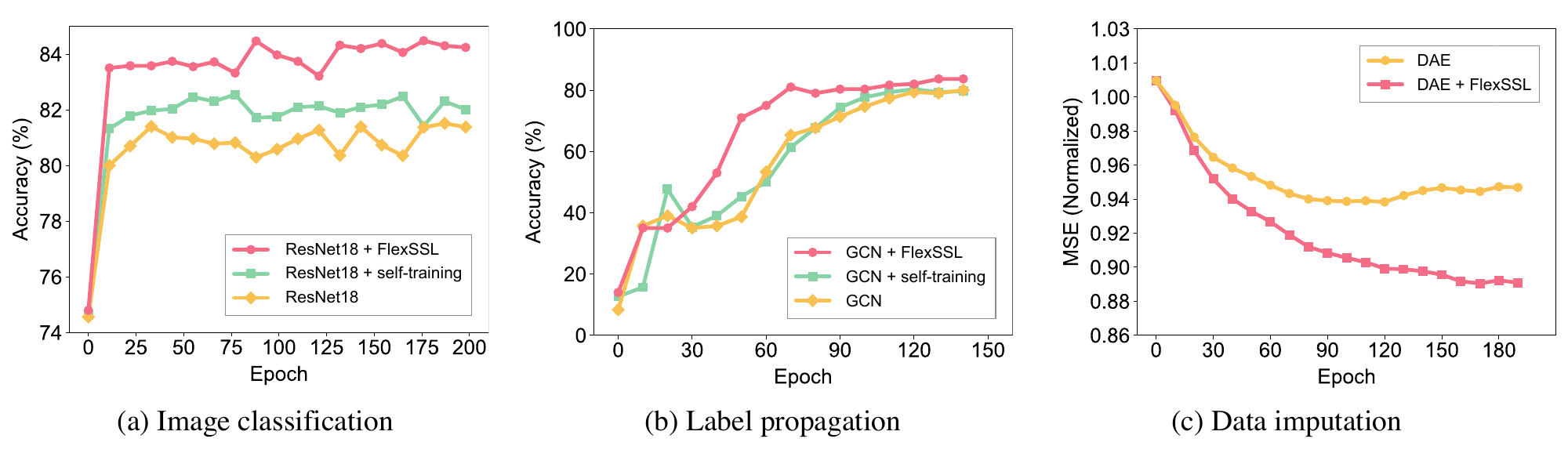}}
    \caption{The convergence of test accuracy under different methods and tasks}
    \label{fig:acc_curver}
%
    \makebox[\textwidth][c]{\includegraphics[width=0.98\textwidth]{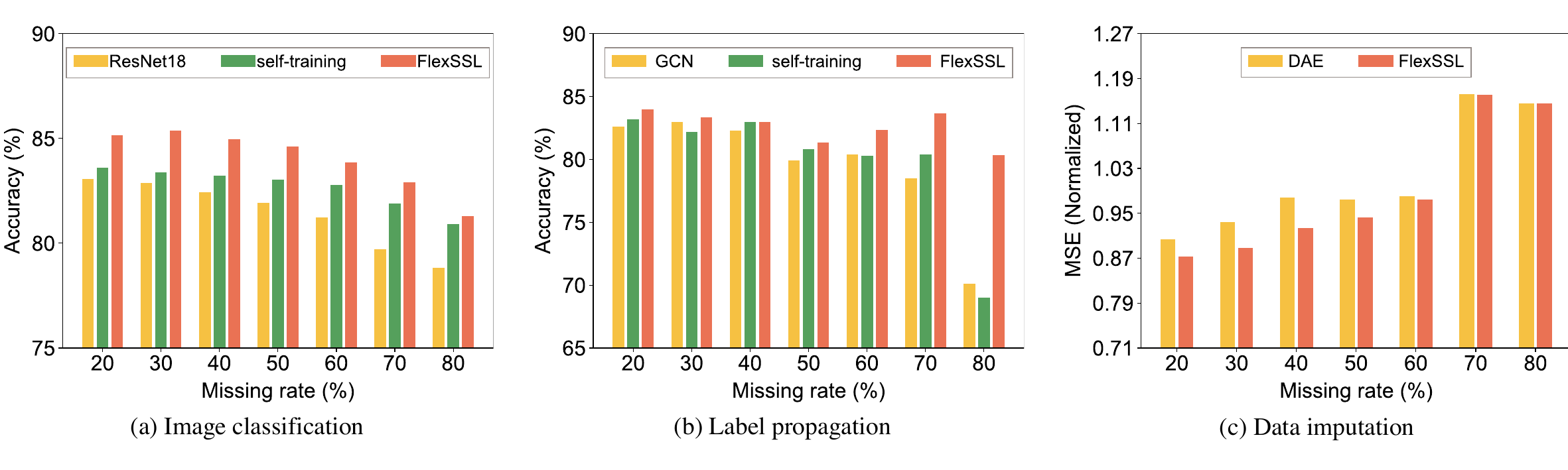}}
    \caption{Performance under different missing rate}
    \label{fig:missing_rate}
  \end{figure*}
   
    \subsubsection{Learning efficiency. } We further compare the performance of FlexSSL with the widely used self-training-based pseudo-labeling method \cite{DBLP:conf/iclr/LaineA17,iscen2019label} in different tasks. Figure \ref{fig:acc_curver} shows the learning curves of image classification, label propagation, and data imputation under missing rates of 40\%, 20\%, and 30\%, respectively. In Figure \ref{fig:acc_curver}, we observe that the learning speed of the main task model with the corrective reweighting loss term $loss_W$ in FlexSSL is significantly faster than the original model. This is primarily due to the use of carefully reweighted soft labels to correct unreliable label information in FlexSSL, as compared to the use of hard pseudo-labels as true label for model learning in self-training.

   \subsubsection{Robustness to missing data. }
   Figure \ref{fig:missing_rate} shows the accuracy of the original, self-trained and FlexSSL-integrated versions of models for tasks with different missing rates. We observe that FlexSSL maintains high accuracy under various missing rates and consistently outperforms the original and self-training versions. This is achieved through the cost-sensitive learning in FlexSSL for different data samples to maximally utilize the hidden information from both labeled and unlabeled data, which lead to more robust and accurate inference. 
   In the label propagation task, although self-training increases the training samples by periodically pseudo-labeling high confidence unlabeled data, the accuracy of the model could decreases due to the involvement of wrong labeling information, which results in subsequent error accumulation and performance drop. FlexSSL, however, maintains high accuracy in spite of high missing rate by continuously soft pseudo-labeling the data to ensure reliable pseudo-labels are properly used to facilitate main task training.
     
   \subsubsection{Robustness to error accumulation. } 
\begin{figure*}[t]
    \centering
    \makebox[\textwidth][c]{\includegraphics[width=0.98\textwidth]{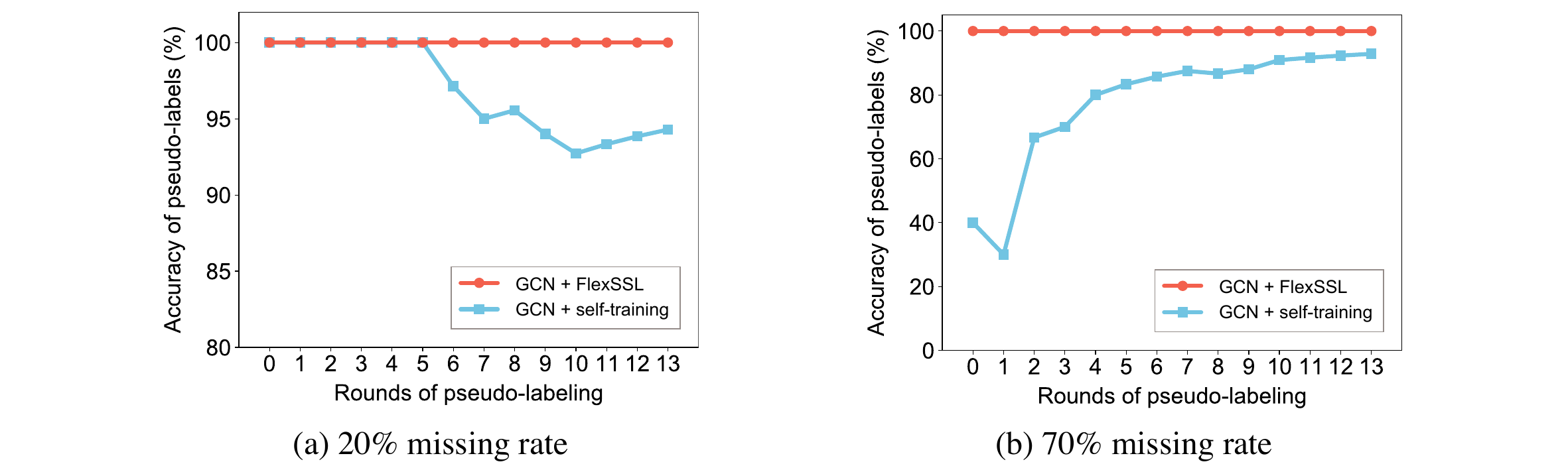}}
    \caption{Impact of Error accumulation on accuracy of pseudo-labels in the label propagation task}
    \label{fig:error_accumulation}
  \end{figure*}

To further investigate the impact of error accumulation in pseudo-labels during model training, we record the accuracy of pseudo-labels after each round of pseudo-labeling in self-training. The test accuracy of these samples in the FlexSSL version of the model are also recorded for comparison.
   To ensure the stability of self-training, we add pseudo-labels every 10 training epochs. Different from self-training that only label high confidence unlabeled samples, FlexSSL introduces pseudo-labels for all unlabeled data $Y_U$ but with soft-labeling weights. 
   Figure \ref{fig:error_accumulation} shows the error accumulation trends
   for self-training and FlexSSL in the label propagation task.
   When the missing rate is low (Figure \ref{fig:error_accumulation}(a)), 
   self-training maintains a high accuracy at the beginning, but model performance will fluctuate and
   decrease with the involvement of more potentially problematic pseudo-labels.
   On the other hand, when the missing rate is high and available information is scarce (Figure \ref{fig:error_accumulation}(b)), gradually involving pseudo-labeled data is beneficial for self-training, resulting in slowly increasing test accuracy, although the model still struggles to achieve high accuracy.
   Despite the occurrence of error accumulation in self-training, we barely observe such phenomenon in FlexSSL. The test accuracy of the pseudo-labels remains high and stable across different rounds of pseudo-labeling, which clearly suggests the ability of FlexSSL in avoiding error accumulation in pseudo-labels.
    

   \subsection{Performance and Efficiency Comparison with SOTA}
   To further validate the performance and efficiency of FlexSSL, we also conduct experiments on the semi-supervised image classification task against advanced SSL methods on CIFAR-10 and CIFAR-100 datasets with 4000 labeled samples. Table \ref{tab:sota} presents the mean and standard deviation results, as well as the training time until convergence of each method. FlexSSL achieves superior performance and efficiency over almost all advanced SSL methods, including MT \cite{DBLP:conf/iclr/TarvainenV17}, UDA \cite{DBLP:conf/nips/XieDHL020}, Mixmatch \cite{DBLP:conf/nips/BerthelotCGPOR19}, FlexMatch \cite{zhang2021flexmatch}. It also achieves comparable results to the SOTA method UPS \cite{DBLP:conf/iclr/RizveDRS21}, but with only 1/10th of the training time.
   Under the same hardware setting, it takes more than 120 hours to train UPS on both CIFAR-10 and CIFAR-100 to achieve 93\% and 58\% accuracy respectively; whereas FlexSSL takes only 10 hours and 2.5 hours to reach final convergence, reducing the training time by at least 92\% to reach the similar level of accuracy.
   
\begin{figure*}[h]
  \centering
  \makebox[\textwidth][c]{\includegraphics[width=0.98\textwidth]{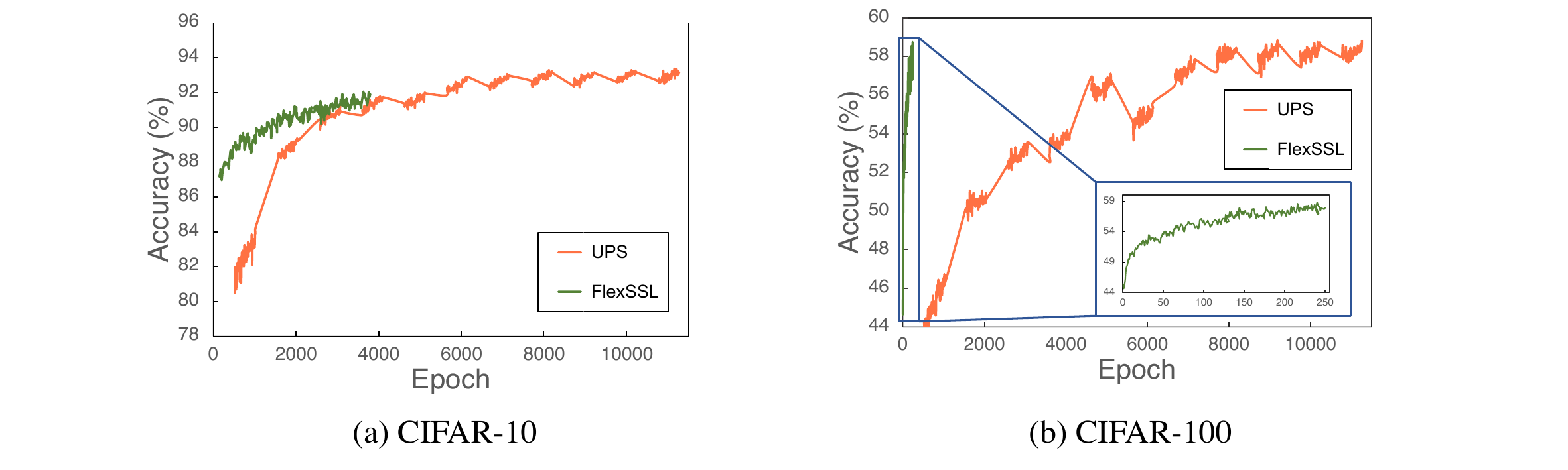}}
  \caption{Learning curves of CCS and UPS on the CIFAR-10 and CIFAR-100 dataset with 4000 labels}
  \label{fig:convergence_with_advanced_ssl}
\end{figure*}

   Meanwhile, we plot the convergence curves of the top-2 methods with the highest accuracy in Table \ref{tab:sota} to compare the efficiency of the training process in Figure \ref{fig:convergence_with_advanced_ssl}.
   UPS employs costly label confidence measures combined with the negative sample learning to select high-quality pseudo-labels to improve model performance, but these also make the algorithm computationally expensive.
   By contrast, FlexSSL does not contain any pseudo-label selection process, but simply pseudo-label all unlabeled samples and use the soft-labeling weights to unevenly update the loss of different samples.
   The great computational efficiency of FlexSSL combined with its generality and applicability to a wide range of SSL tasks, demonstrate its great potential for solving various real-world large-scale problems.

   \begin{table}[t]
      \small
      \centering
      \scalebox{1}{
          \begin{tabular}{l|c|c}
             \toprule
             \multirow{1}{*}{Methods} & \multicolumn{1}{c|}{CIFAR-10 (time)} & \multicolumn{1}{c}{CIFAR-100 (time)} \\
             \midrule
             MT & 11.41 $\pm$ 0.25 (10h)& 45.36 $\pm$ 0.20 (6h)\\ 
             UDA & {8.21} (19h)& {45.07} (18h)\\ 
             Mixmatch & {10.88} (5h)& {45.66} (13h)\\ 
             Flexmatch & {8.08} (17.5h)& {42.26} (5h)\\ 
             UPS & \textbf{6.39 $\pm$ 0.03} (120h)& \textbf{40.77 $\pm$ 0.10} (120h)\\ 
             \midrule
             FlexSSL & \textbf{7.85 $\pm$ 0.03} (10h)& \textbf{41.37 $\pm$ 0.12} (2.5h)\\ 
             
             \bottomrule
          \end{tabular} 
      }
    \caption{Error rate (\%) and model training time on the CIFAR-10 and CIFAR-100 test set with 4000 labels. We bold the top two highest scores.}
      \label{tab:sota}
    \end{table}
    

\subsection{Additional Ablation Study}
\begin{figure*}[h]
  \centering
  \makebox[\textwidth][c]{\includegraphics[width=1.15\textwidth]{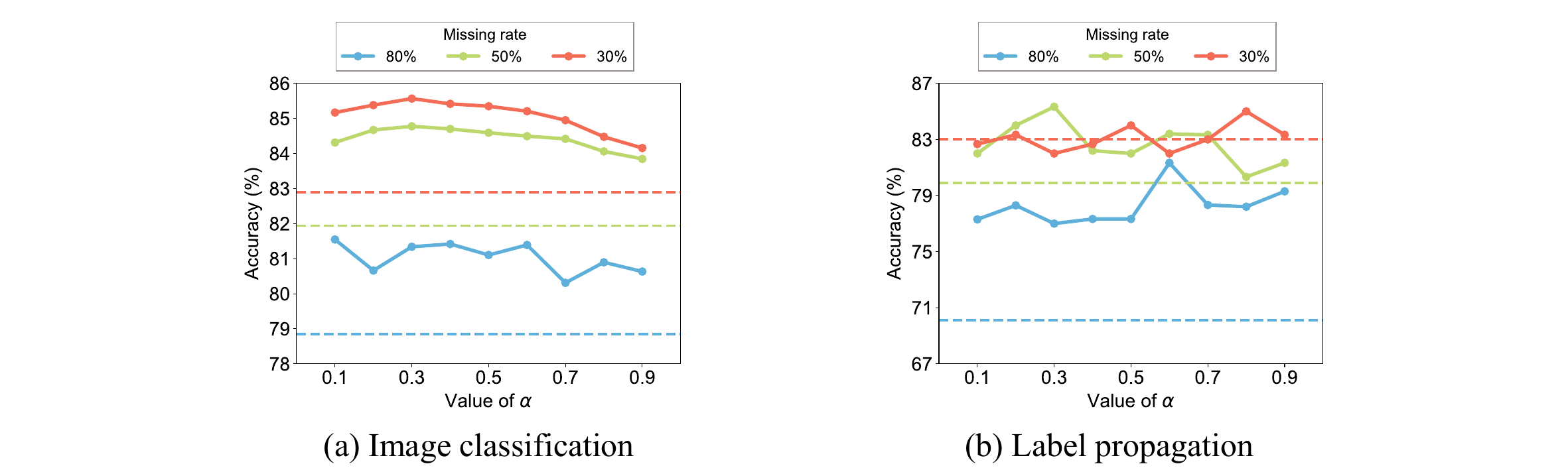}}
  \caption{Test accuracy under different weight parameter $\alpha$}
  \label{fig:alpha}
\end{figure*}

In this section we present the ablation study of the weight parameter $\alpha$ under the image classification task to analyze the impact of $\alpha$ in FlexSSL.
We selected three different levels of missing rates in our experiments - low (30\%), medium (50\%) and high (80\%). Figure \ref{fig:alpha} shows the classification accuracy obtained by FlexSSL using different $\alpha$ values under different missing rates. We observe that a larger $\alpha$ will make the main task favors more on minimizing the reweighting loss rather than simply minimizing the difference between model predicted labels and the given labels $\tilde{Y}$. In our exeriment results in Figure 6, it actually suggests that with the $\alpha$ taking value from the recommended range $[0.1,0.9]$, FlexSSL is not very sensitive to $\alpha$. The maximum difference of accuracy between the worst and best choice of $\alpha$ is only about $2\%$ under different missing rates. The insensitivity to $\alpha$ is another advantage of FlexSSL, as we do not need to perform complex parameter tuning in order to learn a robust and high performance model, which is very implementation friendly.


\subsection{Interpretability of the Discriminator Outputs}
\begin{figure*}[h]
  \centering
  \makebox[\textwidth][c]{\includegraphics[width=0.98\textwidth]{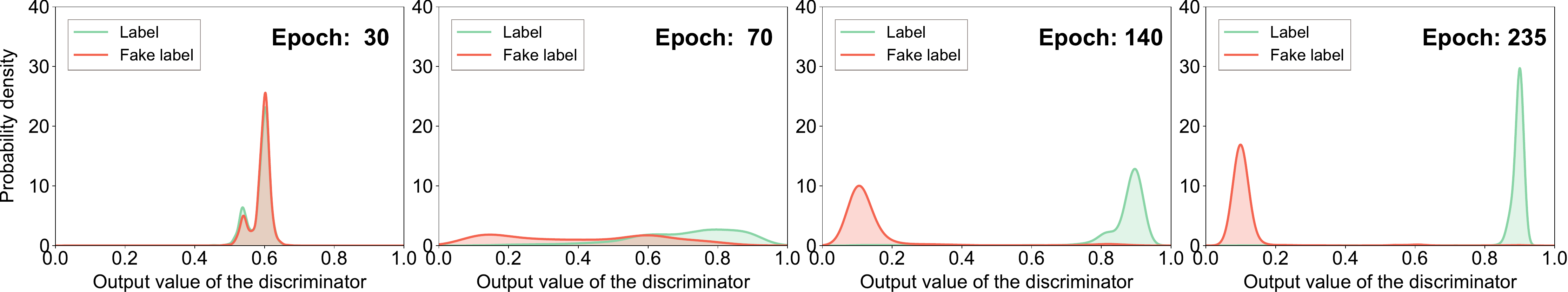}}
  \caption{The probability distribution given by the discriminator during the training process}
  \label{fig:dis_prob}
\end{figure*}

To evaluate the judgement of the discriminator $d$, we analyze its output value distribution on the labeled and intentionally introduced fake/noisy labeled data in different training stages of image classification task. The four plots in Figure \ref{fig:dis_prob} clearly show the transition of an improved discriminator, which gradually gains better capability to discriminate true labels ($p\rightarrow 1$) and fake/noisy labels ($p\rightarrow 0$). The interpretability of the discriminator is a byproduct of FlexSSL, which can be particularly useful for many real-world FlexSSL tasks or scenarios involving error or noise in labeled data.

\section{Related Work}
\subsubsection{Self-training via pseudo-labeling.}
Pseudo-labeling \cite{triguero2015self,yarowsky1995unsupervised,lee2013pseudo} is one of the most widely used self-training method in semi-supervised learning, which uses a supervised model that is iteratively trained on both labeled and pseudo-labeled data in previous iterations of the algorithm. The basic assumption in most algorithms \cite{yarowsky1995unsupervised,lee2013pseudo,DBLP:conf/iclr/LaineA17,tarvainen2017mean,iscen2019label} is that an unlabeled instance can be labeled for training if the underlying model has high confidence in its prediction. When the model makes inaccurate predictions, it is likely to cause error accumulation that negatively impacts training  \cite{cai2013heterogeneous,arazo2020pseudo}. To alleviate this problem, many studies
\cite{DBLP:conf/icdm/ZhouKT12,DBLP:conf/iccv/ZouYLKW19,DBLP:conf/icassp/Kahn0H20}
guide the model learning based on various label confidence measures and avoid the model from being overly confident on incorrect pseudo-labels.
The above works only focus on leveraging both labeled and unlabeled data, but not fully applicable to scenarios with noisy labeled data.
FlexSSL considers the contribution of both labeled and unlabeled data through soft labeling, which better addresses aforementioned drawbacks.

\subsubsection{Learning from noisy labels.}
The lack of high-quality labels is common in many real-world scenarios \cite{DBLP:journals/jmlr/NatarajanDRT17, DBLP:conf/icml/SchnabelSSCJ16}. Erroneous or noisy labels can severely degrade the generalization performance. Learning from noisy labels \cite{DBLP:journals/corr/abs-2007-08199} is becoming an increasingly important task in modern deep learning applications. Early studies \cite{DBLP:journals/tit/Cesa-BianchiSS11} require unbiased gradient estimates of the loss to provide learning guarantees. Some recent studies \cite{DBLP:journals/jmlr/NatarajanDRT17, DBLP:conf/icml/SchnabelSSCJ16,DBLP:conf/aaai/Shen0ZK20, algan2021meta} provide unbiased estimators for evaluating and training under noise data. However, these methods still require that the observed data have a relatively well-behaved (noise-free) underlying distribution, thus the noisy data can be weighted according to the same distribution. Satisfying such requirements is difficult in practice and the observed distribution may be biased across the data. In contrast, FlexSSL bypasses the dependence on the original data distribution and learn label confidence measures adaptive using an additional discriminator to facilitate model training.

\section{Conclusion}
We introduce a novel and efficient semi-supervised learning framework called FlexSSL, which can be easily incorporated into a wide range of SSL models for enhanced performance without the need of any domain-specific knowledge. FlexSSL improves semi-supervised model learning by constructing a cooperative-yet-competitive ``game'' that establishes cooperation between a main self-interested semi-supervised learning task and a companion task of inferring the observability of labels. Meanwhile, it also offers additional confidence measures for the observed and inferred data labels, which can be particularly useful for the practical applications with noisy labels. Through extensive evaluations, we show that FlexSSL can consistently improve the performance and robustness of semi-supervised classification and regression models with great learning efficiency.

%

 \appendices
 \section{Experiment Settings and Implementation Details}
 \label{experiment_setting}
 \subsection{Image classification}
 The base learner ($f$) is a ResNet18 \cite{he2016deep} where the input channels of the first convolutional layer are replaced with 1. We use the Kaiming Normalization \cite{DBLP:conf/iccv/HeZRS15} to initialize the parameters of $f$. All input samples of training set and test set are transformed with the mean and unit standard deviation for normalization. The discriminator consists of two feature extractors for $X$ and $\hat{Y}$, respectively, and a fusion layer whose inputs are the extracted $X$ and $\hat{Y}$ features and the main task information $g$. The feature extractor of $X$ is also a ResNet18 with the number of input channels of the first convolutional layer as 1, and it outputs features with dimension 1. The feature extractor for $\hat{Y}$ is a network with three-layer FC, where the first two layers are each followed by a ReLU layer. We fuse the extracted features of $X$ and $\hat{Y}$ by the Hadamard product. And concatenate the loss $g$ to the fused embeddings of $X$, $\hat{Y}$, then fuse them through a one-layer FC. We feed the output into a sigmoid layer to output the probabilities. We use the Adam optimizer with a learning rate of 0.001 to train the base learner $f$ and discriminator $d$ simultaneously for a total of 300 epochs. We set $\alpha=0.6$ in all experiments for this task.
 
 \subsection{Label propagation}
 We train a two-layer GCN as described in \cite{DBLP:conf/iclr/KipfW17} and evaluate prediction accuracy on a test set of 1,000 labeled examples. We initialize weights using the initialization described in \cite{glorot2010understanding} and accordingly (row-)normalize input feature vectors. For discriminator $d$, we use a five-layer and two-layer FC to encode the input $X$ and $\hat{Y}$, and we fuse the hidden embedding of $X$ and $\hat{Y}$ by Hadamard product. We connect $g$ to the fusion embedding of $X,\hat{Y}$ and fuse it through one-layer FC and then fed to a sigmoid layer to give the final probability. 
 All FC layers is followed by a ReLU operations except the output layer. We train all models for a maximum of 150 epochs (training iterations) using Adam \cite{DBLP:journals/corr/KingmaB14} with a learning rate of 0.01. We set $\alpha=1$ in all experiments for this task.
 
 \subsection{Data imputation}
 The denoise autoencoder we used in the data imputation task as described in \cite{DBLP:conf/pakdd/GondaraW18} consists of a dropout layer and an encoder composed of four-layer FC and a decoder composed of four-layer FC, where the dropout layer has $p=0.1$. For discriminator $d$, we use three two-layer FC to encode the input $X, \hat{Y}$ and $g$, and we fuse the hidden embeddings by two one-layer FC and then fed to a sigmoid layer to give the final probability. 
 All FC layers in the discriminator are followed by ReLU operations, while all FC layers in the DAE are followed by tanh operations, with the exception of the output layer. We train all models for a maximum of 100 epochs using Adam \cite{DBLP:journals/corr/KingmaB14} with a learning rate of $10^{-5}$ and early stopping with a window size of 10, i.e. we stop training if the validation loss does not decrease for 10 consecutive epochs.
 We set $\alpha=1$ in all experiments for this task.
 
 \subsection{Comparison with advanced SSL methods on the image classification task}
 For the comparative experiments with advanced SSL methods, we use the CNN-13 architecture that is commonly used in benchmarking SSL algorithms \cite{DBLP:conf/nips/OliverORCG18} and set $\alpha=0.9$ for FlexSSL in this task. For a fair comparison, we compare against the methods which report results using the same CNN-13 backbone architecture: MT \cite{DBLP:conf/iclr/TarvainenV17}, UDA \cite{DBLP:conf/nips/XieDHL020}, Mixmatch \cite{DBLP:conf/nips/BerthelotCGPOR19}, UPS \cite{DBLP:conf/iclr/RizveDRS21} and Flexmatch \cite{zhang2021flexmatch}. We keep the experiment setting the same as in the UPS paper \cite{DBLP:conf/iclr/RizveDRS21} and Flexmatch paper \cite{zhang2021flexmatch} for comparison. The technical details of above baselines algorithms please refer to their papers.
 For the computational performance evaluation, both UPS and FlexSSL are trained under the same hardware setting: 24-core 2.4GHz CPU and Nvidia A100 GPU.



\ifCLASSOPTIONcompsoc


\ifCLASSOPTIONcaptionsoff
  \newpage
\fi

\bibliographystyle{IEEEtran}
\bibliography{bare_adv}

\begin{IEEEbiography}[{\includegraphics[width=1in,height=1.25in,clip,keepaspectratio]{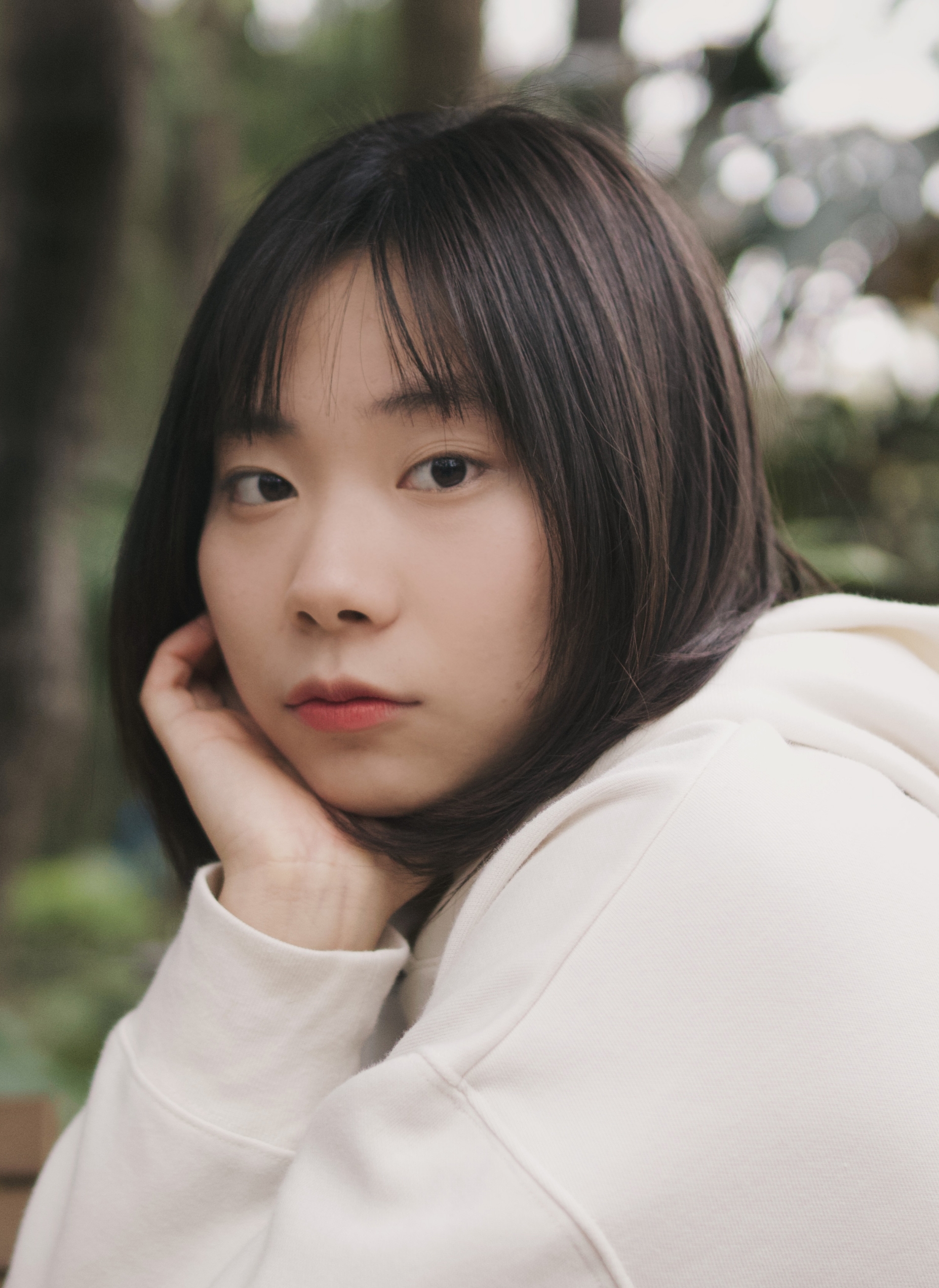}}]{Huiling Qin}
	Huiling Qin is a Ph.D. student in the School of Computer Science and Technology, XiDian University. Before that, she received the B.E degree from XiDian University in 2018. Her current research interest involves urban computing, spatio-temporal data mining, and graph-based model.
\end{IEEEbiography}
  
\begin{IEEEbiography}[{\includegraphics[width=1in,height=1.25in,clip,keepaspectratio]{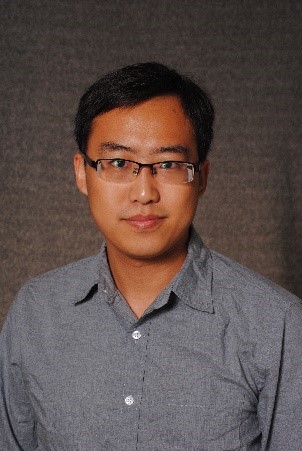}}]{Xianyuan Zhan} 
	Xianyuan Zhan is a research assistant professor at the Institute for AI Industry Research (AIR), Tsinghua University. He received a dual Master’s degree in Computer Science and Transportation Engineering, and a PhD degree in Transportation Engineering from Purdue University. Before joining AIR, Dr. Zhan was a data scientist at JD Technology and also a researcher at Microsoft Research Asia (MSRA). His research interests include deep reinforcement learning, urban computing and big data analytics in transportation.
\end{IEEEbiography} 

\begin{IEEEbiography}[{\includegraphics[width=1in,height=1.25in,clip,keepaspectratio]{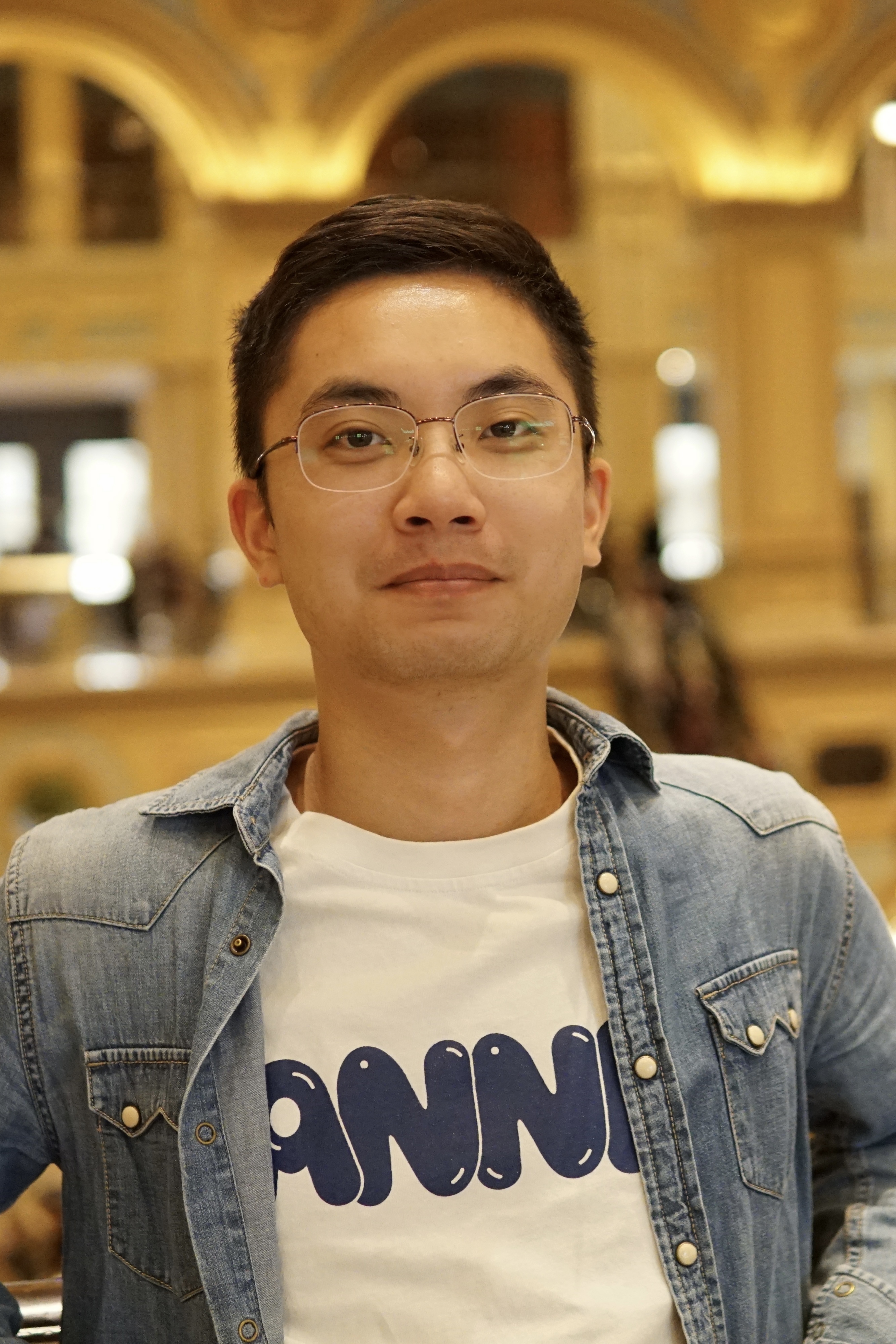}}]{Yuanxun Li}
	Yuanxun Li is a engineer in Kingsoft. Before that, he received the M.S degree from Sun Yat-sen University in 2021 and received the B.E degree from Xi'an Jiaotong University in 2018. He current research interest involves anomaly detection, data mining, and AIOPS.
\end{IEEEbiography}

\begin{IEEEbiography}[{\includegraphics[width=1in,height=1.25in,clip,keepaspectratio]{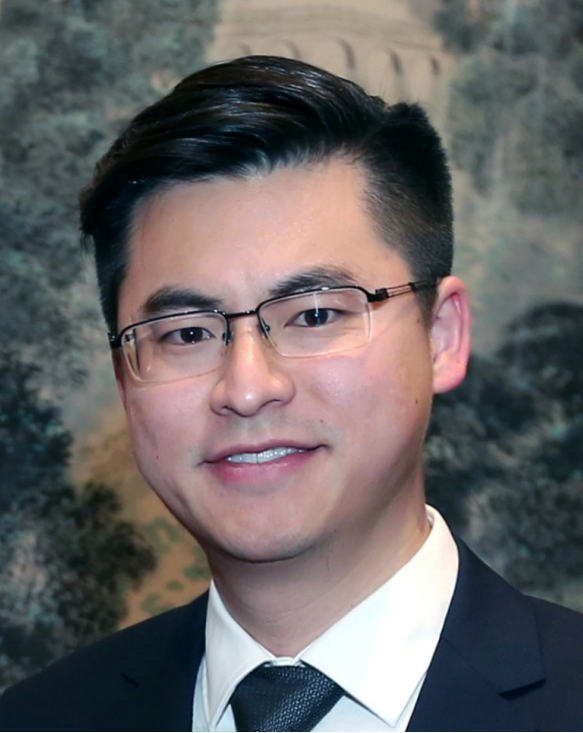}}]{Yu Zheng}
    Dr. Yu Zheng is a Vice President of JD.COM and the Chief Data Scientist of JD Technology. He also leads the Intelligent Cities Business Unit as the president and serves as the managing director of JD Intelligent Cities Research. His research interests include big data analytics, spatio-temporal data mining, machine learning, and artificial intelligence. Before joining JD Technology, he was a senior research manager at Microsoft Research and is also a Chair Professor at Shanghai Jiao Tong University, an Adjunct Professor at Hong Kong University of Science and Technology. Zheng was named an ACM Distinguished Scientist in 2014 and elevated to an IEEE Fellow in 2020 for his contributions to spatio-temporal data mining and urban computing.
\end{IEEEbiography}

\end{document}